\documentclass{llncs}
\usepackage{makeidx}  
\usepackage{amssymb}
\usepackage{mathtools}
\usepackage{graphicx}
\usepackage{footnote}
\usepackage{url}
\usepackage{float}
\usepackage{todonotes}
\usepackage{booktabs}
\usepackage{mdframed}
\usepackage{graphicx}
\usepackage{multirow}
\usepackage{multicol}
\usepackage{tabularx}
\usepackage{xspace}
\usepackage{comment}
\usepackage{tikz}
\usetikzlibrary{shapes.gates.logic.US,trees,positioning,arrows,shapes.geometric}
\usepackage{todonotes}
\usepackage{lscape}
\usepackage{tabularx}
\newcommand{\eval}[1]{[[#1]]}
\newcommand{\EAGLE}{\textsc{Eagle}\xspace}
\newcommand{\CANONICAL}{\textsc{Canonical}\xspace}
\newcommand{\HELIOS}{\textsc{Helios}\xspace}

\newcommand{\LIMES}{\textsc{Limes}\xspace}
\usepackage{llncsdoc}

\begin{document}

\title{An Evaluation of Models for Runtime Approximation in Link Discovery}
\author{Kleanthi Georgala\inst{1} \and Micheal Hoffmann\inst{1} \and Axel-Cyrille Ngonga Ngomo\inst{1}}
\institute{AKSW Research Group, University of Leipzig, Germany
\email{\{georgala|mhoffmann|ngonga\}@informatik.uni-leipzig.de}}
\maketitle
\begin{abstract}

Time-efficient link discovery is of central importance to implement the vision of the Semantic Web. Some of the most rapid Link Discovery approaches rely internally on planning to execute link specifications. In newer works, linear models have been used to estimate the runtime the fastest planners. However, no other category of models has been studied for this purpose so far. In this paper, we study non-linear runtime estimation functions for runtime estimation. In particular, we study exponential and mixed models for the estimation of the runtimes of planners. To this end, we evaluate three different models for runtime on six datasets using 400 link specifications. We show that exponential and mixed models achieve better fits when trained but are only to be preferred in some cases. Our evaluation also shows that the use of better runtime approximation models has a positive impact on the overall execution of link specifications.
\end{abstract}

\section{Introduction}
\label{sec:introduction}
Link discovery frameworks are of utmost importance during the creation of Linked Data~\cite{auer2013introduction}. This is due to their being the key towards the implementation of the fourth Linked Data principle, i.e., the provision of links between datasets.\footnote{\url{https://www.w3.org/DesignIssues/LinkedData.html}}
Two main challenges need to be addressed by Link Discovery frameworks~\cite{ngomo2012link,helios}.
First, they need to address the \emph{accuracy challenge}, i.e., they need to generate correct links. A plethora of approaches have been developed for this purpose and contain algorithms ranging from genetic programming to probabilistic models. In addition to addressing the need for accurate links, link discovery frameworks need to address the \emph{challenge of time efficiency}. This challenge comes about because of the mere size of knowledge bases that need to be linked. In particular, large knowledge bases such as LinkedTCGA~\cite{saleem2014big} contain more than 20 billion triples. 

One of the approaches to improving the scalability of link discovery frameworks is to use planning algorithms in a manner akin (but not equivalent to) their use in databases~\cite{helios}. In general, planners rely on cost functions to estimate the runtime of particular portions of link specifications.
So far, it has been assumed that this cost function is linear in the parameters of the planning, i.e., in the size of the datasets and the similarity threshold. However, this assumption has never been verified.
In this paper, we address exactly this research gap and \emph{study how well other models for runtime approximation perform}. In particular, we study linear, exponential and mixed models for runtime estimation. The contributions of this paper are thus as follows:
\begin{itemize}
	\item We present three different models for runtime approximation in planning for Link Discovery.
	\item We compare these models on six different datasets and study how well they can approximate runtimes of specifications as well as with respect to how well they generalize across datasets. 
	\item We integrate the models with the \textsc{Helios} planner for Link Discovery as described in~\cite{helios} and compare their performance using 400 specifications.
\end{itemize}

The rest of the paper is structured as follows: In Section~\ref{sec:preliminaries}, we present the concept and notations necessary to understand this work. The subsequent section, Section~\ref{sec:approach}, presents the runtime approximation problem and how it can be addressed by different models. We then delve into a thorough evaluation of these models in Section~\ref{sec:evaluation} and compare the expected runtimes generated by the models at hand with the real runtimes of the Link Discovery framework. We also study the transferability of the results we achieve and their performance when planning whole link specifications. Finally, we recapitulate our results and conclude.

\section{Preliminaries}
\label{sec:preliminaries}
In this section, we present the necessary concepts and notations to understand the rest of the paper. We begin by giving a description of a knowledge base $K$ and Link Discovery (LD), we continue by providing a formal definition of a link specification (LS) and its semantics and we finish our preliminary section with an explanatory presentation of a plan, its components and its relation to a LS.

\paragraph{Knowledge Base.} A knowledge base $K$ is a set of triples $(s, p, o) \in (\mathcal{R} \cup \mathcal{B}) \times \mathcal{P} \times (\mathcal{R} \cup \mathcal{B} \cup \mathcal{L})$, where $\mathcal{R}$ is the set of all RDF resources, $\mathcal{P} \subseteq \mathcal{R}$ is the set of all RDF properties, $\mathcal{B}$ is the set of all RDF blank nodes and $\mathcal{L}$ is the set of all literals.

\paragraph{Link Discovery.} Given two (not necessarily distinct) sets of RDF resources $S$ and $T$ and a relation $R$ (e.g, \texttt{directorOf}, \texttt{owl:sameAs}), the main goal of LD is to discover the set (\emph{mapping}) $\{(s,t) \in S \times T: R(s,t)\}$. Given that this task can be very tedious (especially when $S$ and $T$ are large), LD frameworks are commonly used to achieve this computation.

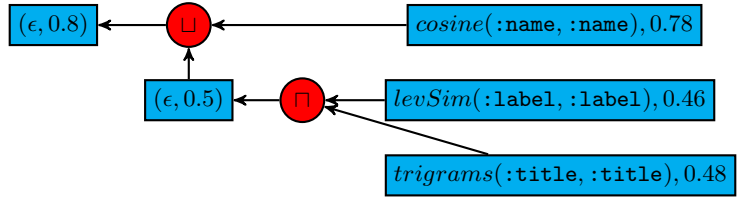
\begin{figure}[htb]
	\centering
\begin{tikzpicture}[scale = 1, every node/.style={scale=1}]
	  \begin{scope}[thick,
			node distance=1.0cm,on grid,>=stealth',
			filter/.style={rectangle,draw,fill=cyan},
			op/.style={circle,draw,fill=red}]
			
	    \node [filter](filter1)[]{$(\epsilon, 0.8)$};
		\node [op](intersection)	[right =of filter1,xshift=0.8cm ]	{$\sqcup$} edge [->] (filter1);	    
	    \node [filter] (label3)[right =of intersection.west,anchor=west,xshift=2.2cm]	{$cosine(\texttt{:name},\texttt{:name}), 0.78$} edge [->] (intersection) ;
	    	\node [filter](filter2)[below =of intersection]{$(\epsilon, 0.5)$}edge [->] (intersection);
	    \node [op] (intersection2)[right =of filter2.west,xshift=.8cm]	{$\sqcap$}  edge [->] (filter2);
	    	\node [filter] (label2)[right =of intersection2.west,xshift=.4cm]	{$levSim(\texttt{:label},\texttt{:label}), 0.46$}edge [->] (intersection2) ; 
		\node [filter] (label1)[below =of label2.west,anchor=west]{$trigrams(\texttt{:title}, \texttt{:title}), 0.48$}edge [->] (intersection2);
		
	  \end{scope}
	\end{tikzpicture}
	\caption{Graphical representation of an example LS}
	\label{fig:specexample}
\end{figure}

\paragraph{Link Specification.} Declarative LD frameworks use link specifications (LSs) to describe the conditions for which $R(s,t)$ holds for a pair $(s, t) \in S \times T$. A LS consists of two basic components:
\begin{itemize}
	\item \emph{similarity measures} which allow the comparison of property values of resources found in the input data sets $S$ and $T$.  We define an \emph{atomic similarity measure} $m \in M$ as a function $m:S \times T \times \mathcal{P}^2 \rightarrow [0, 1]$.  We write $m(s, t, p_{s},p_{t})$ to signify the similarity of $s$ and $t$ w.r.t. their properties $p_{s}$ resp. $p_{t}$. 
	\item \emph{operators} $op \in \{\sqcup, \sqcap, \backslash\}$ that allow the combination of two \emph{similarity measures}. 
\end{itemize}
An atomic LS consists of one similarity measure and has the form $(m(p_s, p_t), \theta)$ where $\theta \in [0,1]$.  A complex LS $L = op(L_1, L_2)$ consists of two LS, $L_1$ and $L_2$. We call $L_1$ the \emph{left sub-specification} and $L_2$ the \emph{right sub-specification} of $L$. We denote the semantics (i.e., the results of a LS for given sets of resources $S$ and $T$) of a LS $L$ as $\eval L$ and call it a mapping. We begin by assuming the natural semantics of the combinations of measures. Filters are pairs $(f, \tau)$, where (1) $f$ is either empty (denoted $\epsilon$) or a combination of similarity measures by means of specification operators and (2) $\tau$ is a threshold. Note that an atomic specification can be regarded as a filter $(f, \tau, X)$ with $[[X]] = S \times T$. We will thus use the same graphical representation for filters and atomic specifications. We call $(f, \tau)$ the \emph{filter of $L$} and denote it with $\varphi(L)$. For our example $L$ in Fig.~\ref{fig:specexample}, $\varphi(L) = (\epsilon, 0.7)$. 
We denote the \emph{operator of a LS} $L$ with $op(L)$. For $L = (f, \tau, \omega(L_1, L_2))$, $op(L) = \omega$. The operator of the LS shown in our example is  $\sqcup$. The semantics of LSs are then as shown in Table~\ref{tab:semantics}. 
\paragraph{Execution Plan.} To compute the mapping $\eval L$ (which corresponds to the output of $L$ for a given pair ($S, T$)), LD frameworks implement (at least partly) a generic architecture consisting of a rewriter (optional), a planner (optional) and an execution engine (necessary).
The \emph{rewriter} performs algebraic operations to transform the input LS $L$ into a LS $L'$ (with $\eval L = \eval L'$) that is potentially faster to execute.  
The most common planner is the \emph{canonical planner} (dubbed \CANONICAL), which simply traverses $L$ in post-order and has its results computed in that order by the execution engine.\footnote{Note that the planner and engine are not necessarily distinct in existing implementations.}  For the LS shown in Fig.~\ref{fig:specexample}, the execution plan returned by \CANONICAL would thus foresee to first compute the mapping $M_1 = [[(\texttt{trigrams}(\texttt{:title},\texttt{:title}), 0.48)]]$ of pairs of resources whose property \texttt{title} has a cosine similarity greater or equal to $0.48$. The computation of $M_2 = [[(\texttt{levenSim}(\texttt{:label},\texttt{:label}), 0.46)]]$ would follow. Step 3 would be to compute $M_3 = M_1 \sqcap M_2$ while abiding by the semantics described in Table~\ref{tab:semantics}. Step 4 would be to obtain $M_4$ by filtering the results and keeping only the pairs that have a similarity above 0.5. Step 5 would be $M_5 = [[(\texttt{cosine}(\texttt{:name},\texttt{:name}), 0.78)]]$ and Step 6 would be to compute $M_6 = M_4 \sqcup M_5$. Finally, Step 7 would be to filter out the pairs of links in $M_6$ that have a similarity less than 0.8.
Given that there is a 1-1 correspondence between LS and the plan generated by the canonical planner, we will reuse the representation of LS devised above for plans. The sequence of steps for such a plan is then to be understood as the sequence of steps that would be derived by \CANONICAL for the LS displayed.

\begin{table}[htb]
\caption{Semantics of link specifications}\label{tab:semantics}
	\centering
	\begin{tabular}{cl}
		\toprule
		$L$					&	$[[L]]$ \\\midrule
		$(m, \theta)$		&	$\{(s, t, m(s, t)) \in S \times T: m(s, t) \geq \theta\}$ \\
		$(f, \tau, X)$		&	$\begin{cases}
		\{(s, t, r) \in [[X]]: r \geq \tau\} \mbox{ if } f = \epsilon\\
		\{(s, t, r) \in [[X]]: f(s, t) \geq \tau\} \mbox{ else.} 
		\end{cases}$\\
		$\sqcap(L_1, L_2)$ & $\{ (s,t,r) \mid (s,t,r_1) \in [[L_1]] \wedge (s,t,r_2) \in [[L_2]] \wedge r = \min(r_1,r_2)\}$ \\
		$\sqcup(L_1, L_2)$ & $\left \{ (s,t,r) \mid \begin{cases}
			r = r_1 \mbox{ if } \exists (s,t,r_1) \in [[L_1]] \wedge \neg (\exists r_2: (s, t, r_2) \in [[L_2]]),  \\
			r = r_2 \mbox{ if } \exists (s,t,r_2) \in [[L_2]] \wedge \neg (\exists r_1: (s, t, r_1) \in [[L_1]]),\\
			r = \max(r_1,r_2) \mbox{ if }  (s,t,r_1) \in [[L_1]] \wedge (s,t,r_2) \in [[L_2]].  
			\end{cases} \right\}$ \\
		$\backslash(L_1, L_2)$	& $\{ (s,t,r) \mid (s,t,r) \in [[L_1]] \wedge \neg \exists r': (s,t,r') \in [[L_2]] \}$ \\
		$\emptyset(L)$	&	$[[L]]$ \\
		\bottomrule
	\end{tabular}
\end{table}

\section{Runtime Estimation}
\label{sec:approach}
In general, planners aims to estimate the cost of the leaves of a plan, i.e., the runtime of atomic link specifications. So far, linear models \cite{helios} have been used for this purpose but the appropriateness of other models has never been evaluated.
Hence, in this work, we compare non-linear models with linear models to approximate the runtime of of atomic link specifications. Like in previous works, we follow a \emph{sampling-based approach}. First, given a particular similarity measure $m$ (e.g., Levenshtein) and an implementation of the said measure (e.g., \emph{Ed-Join} \cite{xiao2008ed}), we begin by collecting sample of runtimes for a given measure with varying values of $|S|$, $|T|$ and $\theta$.\footnote{We also experimented with the number of trigrams contained in $S$ and $T$ but found that they do not affect the models we considered. An exploration of other parameters remains future work.} These samples can be regarded as the output of a function that can predict the runtime of the implementation of $m$ for which we were given samples. The major question that is to be answered is hence \emph{what is the shape of the runtime evaluation function}?

We tried fitting functions of different shapes to the previously measured runtimes in order to compare their performance when planning the execution of link specifications. 
Formally, these functions are mappings $\phi : \mathbb{N} \times \mathbb{N} \times (0,1] \mapsto \mathbb{R}$, whose value at $(|S|, |T|, \theta)$ is an approximation of the runtime for the link specification with these parameters. 
If $\vec{R} = (R_1, \dots, R_n)$ are the measured runtimes for the parameters $\vec{S} = (|S_1|, \dots, |S_n|)$, $\vec{T} = (|T_1|, \dots, |T_n|)$ and $\vec{\theta} = (\theta_1, \dots, \theta_n)$, then we constrain the mapping $\phi$ to be a local minimum of the L2-Loss: \begin{equation}
E(\vec{S}, \vec{T}, \vec{\theta}, \vec{r}) := \|\vec{R} - \phi(\vec{S}, \vec{T}, \vec{\theta})\|^2,
\end{equation}
writing $\phi(\vec{S}, \vec{T}, \vec{\theta}) = (\phi(|S_1|, |T_1|, \theta_1), \dots, \phi(|S_n|, |T_n|, \theta_n))$.

Within this paper, we consider the following parametrized families of functions: \begin{eqnarray}
\phi_1(S, T, \theta) &=& a + b|S|  + c|T| + d\theta \\
\phi_2(S, T, \theta) &=& \exp{(a + b|S|  + c|T| + d\theta + e\theta^2 )} \\
\phi_3(S, T, \theta) &=& a + (b + c|S| + d|T| + e|S||T|)\exp{(f\theta + g\theta^2) }
\end{eqnarray}
The parameters are then determined by 
\begin{equation}
a^*, b^*, \dots = \text{arg}\min E(\vec{S},\vec{T},\vec{\theta},\vec{R})(a,b, \dots)
\end{equation}
for some local minimum. In the case of $\phi_1$ and $\phi_2$ this problem is linear in nature and we solved it using the pseudo-inverse of the associated Vandermonde matrix. For $\phi_3$ we used 
the Levenberg-Marquardt Algorithm \cite{more1978levenberg} for nonlinear least squares problems, using $1$ as initial guess for all parameters. 

We chose $\phi_1$ as the baseline linear fit. $\phi_2$ is the standard log-linear fit, except for the $\theta^2$ term. We included this term during a grid search for polynomials to perform a log-polynomial fit. Higher orders of $|S|$ or $|T|$ or $\theta$ did not contribute to a better fit. $\phi_3$ can be interpreted as an interpolation of $\phi_1$ and $\phi_2$ with a constant offset $a$.

To exemplify our approach for $\phi_2$, assume we have measured $\vec{S} = ( 458, 458,$ $ 358, 58 ) , \vec{T} = (512, 404, 317, 512 )$ and $\vec{\theta} = (0.5, 0.9, 0.6, 0.7) $. Inserting into eq. (1) and taking the logarithm,
one arrives at the optimization problem
\[
\min_{a,b,c,d,e}
\|
\begin{pmatrix} 1 & 458 & 512  & 0.5 & 0.5^2 \\ 
1 & 458 & 404 & 0.9 & 0.9^2 \\
1 & 358 & 317 & 0.6 & 0.6^2 \\
1 & 58 & 512 & 0.7 & 0.7^2 \\
\end{pmatrix}
\begin{pmatrix}
a \\ b \\ c \\ d \\ e
\end{pmatrix}
-
\begin{pmatrix}
\log(67) \\ \log(4)\\ \log(4) \\ \log(1)
\end{pmatrix}
\|^2
\]
The solution to this least squares problem also is the unique solution of its normal equations:
\[
\begin{pmatrix} 1 & 1 & 1 & 1 \\
458 & 458 & 358 & 58  \\
512 & 404 & 317 & 512 \\
0.5 & 0.9 & 0.6 & 0.7 \\
0.5^2 & 0.9^2 & 0.6^2 & 0.7^2 
\end{pmatrix}
\begin{pmatrix} 1 & 458 & 512  & 0.5 & 0.5^2 \\ 
1 & 458 & 404 & 0.9 & 0.9^2 \\
1 & 358 & 317 & 0.6 & 0.6^2 \\
1 & 58 & 512 & 0.7 & 0.7^2 \\
\end{pmatrix}
\begin{pmatrix}
a \\ b \\ c \\ d \\ e
\end{pmatrix}
\]
\[
= 
\begin{pmatrix} 1 & 1 & 1 & 1 \\
458 & 458 & 358 & 58 \\
512 & 404 & 317 & 512 \\
0.5 & 0.9 & 0.6 & 0.7 \\
0.5^2 & 0.9^2 & 0.6^2 & 0.7^2
\end{pmatrix}
\begin{pmatrix}
\log(67) \\ \log(4)\\ \log(4) \\ \log(1)
\end{pmatrix}
\]
By multiplying and inverting matrices, we arrive at the linear equation
\[
\begin{pmatrix}
a \\ b \\ c \\ d \\ e
\end{pmatrix}
=
\begin{pmatrix} 1 & 458 & 512  & 0.5 & 0.5^2 \\ 
1 & 458 & 404 & 0.9 & 0.9^2 \\
1 & 358 & 317 & 0.6 & 0.6^2 \\
1 & 58 & 512 & 0.7 & 0.7^2 \\
\end{pmatrix}^{+}
\begin{pmatrix}
\log(67) \\ \log(4)\\ \log(4) \\ 0
\end{pmatrix},
\]
where $A^+$ denotes the Moore-Penrose pseudo inverse of $A$ \cite{courrieu2008fast}. Multiplying the matrices, we arrive at 
\[
\begin{pmatrix}
a \\ b \\ c \\ d \\ e
\end{pmatrix}
=
\begin{pmatrix}
-1.028 \\ 0.009 \\ 0.010 \\ 9.821 \\ -9.053
\end{pmatrix}.
\]
Thus we have found the coefficients of the fit function.

\section{Evaluation}
\label{sec:evaluation}
\subsection{Experimental Setup}
We evaluated the three runtime estimation models using six data sets. The first three are the benchmark data sets for LD dubbed Amazon-Google Products, DBLP-ACM and DBLP-Scholar described in \cite{KopckeTR10}.  We also created two larger additional data sets (MOVIES and VILLAGES, see Table~\ref{tab:datasetsproperties}) from the data sets DBpedia, LinkedGeodata and LinkedMDB. \footnote{\url{http://www.linkedmdb.org/}} \footnote{The new data sets as well as a description of how they were constructed are available at \url{http://titan.informatik.uni-leipzig.de/kgeorgala/DATA/}.} The sixth dataset was the set of all English labels from DBpedia 2014. Table~\ref{tab:datasetsproperties} describes the characteristics of the datasets and presents the properties used when linking the retrieved resources for the first four datasets. The mapping properties were provided to the link discovery algorithms underlying our results.

Each of our experiments consisted of two phases: During the \emph{training} phase, we trained each of the models independently. For each model, we computed the set of coefficients for each of the approximation models that minimized the root mean squared error (RMSE) on the training data provided. The aim of the subsequent \emph{test} phase was to evaluate the accuracy of the runtime estimation provided by each model and the performance of the currently best LD planner, \HELIOS \cite{helios}, when it relied of each of the three  models for runtime approximations. Throughout our experiments, we used the algorithms \emph{Ed-Join} \cite{edjoin} (which implements the Levenshtein string distance) and \emph{PPJoin+} \cite{ppjoin} (which implements the Jaccard, Overlap, Cosine and Trigrams string similarity measures) to execute atomics specifications. As thresholds $\theta$ we used random values between $0.5$ and $1$. 

The aim of our evaluation was to answer the following set of questions regarding the performance of the three models \emph{exp}, \emph{linear} and \emph{mixed}:
\begin{itemize}

	\item $Q_1$: How do our models fit each class separately? To answer this question, we began by splitting the source and target data of each of our datasets into two non-overlapping parts of equal size. We used the first half of each source and each target for training and the second half for testing. 
	\begin{itemize}
		\item \emph{Training}: 	We trained the three models on each dataset. For each model, dataset and mapper,  we a) selected 15 source and 15 target random samples of random sizes from the first half of a dataset (Amazon-Google Products, DBLP-ACM, DBLP-Scholar, MOVIES and VILLAGES) and b) compared each source sample with each target sample 3 times. Note that we used the same samples across all models for the sake of fairness. Overall, we ran 675 training experiments to train each model on each dataset. 
		\item \emph{Testing}: To test the accuracy of each model, we ran the corresponding algorithm (\emph{Ed-Join} and \emph{PPJoin+}) with a random threshold between $0.5$ and $1$ and recorded the real runtime of the approach and the runtimes predicted by our three models. Each approach was executed 100 times against the whole of the second half of the same dataset.  		
	\end{itemize}
	
	\item $Q_2$: How do our models generalize across classes, i.e., can a model trained on data from one class be used to predict runtimes accurately on another class?  
\begin{itemize}
	\item \emph{Training}: We trained each model in the same manner as for $Q_1$ on exactly the same five datasets with the sole difference that the samples were selected randomly from the whole dataset.
	\item \emph{Testing}:  Like in the previous series of experiments, we ran \emph{Ed-Join} and \emph{PPJoin+} with a random threshold between $0.5$ and $1$. Each of the algorithms was executed 100 times against the remaining four datasets. 
	\end{itemize}
	
	\item $Q_3$: How do our models perform when trained on a large dataset?
	\begin{itemize}
	\item \emph{Training}: We trained in the same fashion as to answer $Q_1$ with the sole differences that (1)  we used 15 source and 15 target random samples of various sizes between $10,000$ and $100,000$ from (2) the English labels of DBpedia to train our model.
	\item \emph{Testing}: We learned 100 LSs for the Amazon-GP, DBLP-ACM, MOVIES and VILLAGES datasets using the unsupervised version of the EAGLE algorithm \cite{NGLY12}. We chose this algorithm because it was shown to generate meaningful specifications that return high-quality links in previous works. For each dataset, we ran the set of 100 specifications learned by EAGLE on the given dataset by using each of the models during the execution in combination with the HELIOS planning algorithm \cite{helios}, which was shown to outperforms the canonical planner w.r.t. runtime while producing exactly the same results. 
	\end{itemize}
\end{itemize}

Throughout our experiments, we configured \EAGLE by setting the number of generations and population size to 20, mutation and crossover rates were set to 0.6. All experiments for all implementations were carried out on the same 20-core Linux Server running \textit{OpenJDK} 64-Bit Server $1.8.0\_74$ on Ubuntu 14.04.4 LTS on Intel(R) Xeon(R) CPU E5-2650 v3 processors clocked at 2.30GHz. Each \emph{train} experiment and each \emph{test} experiment for $Q_3$ was repeated three times. As evaluation measure, we computed root mean square error (\emph{RMSE}) between the \emph{expected} runtime and the average \emph{execution} runtime required to run each LS. We report all three numbers for each model and dataset.


\begin{table}[H]
\centering
\caption{Entity matching characteristics of data sets}
\label{tab:datasetsproperties}
\resizebox{1.0\textwidth}{!}
{
\begin{tabular}{*{6}{c}} \hline
    \textbf{Data set}           & \textbf{Source (S)} & \textbf{Target (T)} & \textbf{$|S| \times |T|$} & \textbf{Source Property} & \textbf{Target Property}                           \\ \hline
    
Amazon-GP  & Amazon & Google & $4.40 \times 10^6$ & product name, description               & product name, description\\ 
&& Products && manufacturer, price & manufacturer, price\\ \hline
                                
DBLP-ACM  & ACM & DBLP & $6.00 \times 10^6$ & title, authors 
& title  authors \\                          
&& venue && year, venue & year \\ \hline
                                
DBLP-Scholar& DBLP & Google  & $ 0.17 \times 10^9$ & title, authors & title, authors                           \\
&& Scholar && venue, year  & venue, year\\ \hline

MOVIES   & DBpedia & LinkedMDB & $0.17 \times 10^9$ & dbp:name  & dc2:title  \\
 &&&& dbo:director/dbp:name & movie:director/movie:director\_name \\
 &&&& dbo:producer/dbp:name & movie:producer/movie:producer\_name \\
 &&&& dbp:writer/dbp:name & movie:writer/movie:writer\_name \\
 &&&& rdfs:label & rdfs:label \\ 
\hline
   
VILLAGES & DBpedia & LGD & $6.88 \times 10^9$  & rdfs:label & rdfs:label\\
&&&& dbo:populationTotal& lgdo:population \\
&&&& geo:geometry & geom:geometry/agc:asWKT\\ \hline
\end{tabular}
}
\end{table}

\subsection{Results}
To address $Q_1$, we evaluated the performance of our models when trained and tested on the same class. We present the results of this series of experiments in Table~\ref{tab:exp1}. 
For \emph{PPJoin+} (in particular the \emph{trigrams} measure), the \emph{mixed} model achieved the lowest error when tested upon Amazon-GP and DBLP-Scholar, whereas the \emph{linear} model was able to approximate the expected runtime with higher accuracy on the MOVIES and VILLAGES datasets. On average, \emph{linear} model was able to achieve a lower \emph{RMSE} compared to the other two models. For the \emph{Ed-Join}, the \emph{mixed} model outperformed \emph{linear} and \emph{exp} in the majority of datasets (DBLP-Scholar, MOVIES and VILLAGES) and obtained the lowest \emph{RMSE} on average. As we observe in Table~\ref{tab:exp1}, for both measures, the \emph{exp} model retrieved the highest error on average and is thus the model less suitable for runtime approximations. Especially, for the \emph{Ed-Join}, \emph{exp} had the worst performance in four out of the five datasets and retrieved the highest \emph{RMSE} among the different test datasets for VILLAGES.  This clearly answers our first questions: the \emph{linear} and \emph{mixed} approximation models are able achieve the smallest error when trained on the class on which they are tested.

\begin{table}[htb]
\centering
\caption{Average expected runtime, average execution time and root mean square error for the first five datasets for training and testing on the same class. All runtimes are presented in milliseconds.}
\label{tab:exp1}
\resizebox{1.0\textwidth}{!}
{
\begin{tabular}{rr|rrr|rrr|rrr} \hline

\textbf{Measures}& \textbf{Model}& \multicolumn{3}{r}{\textbf{Amazon-GP}} & \multicolumn{3}{r}{\textbf{DBLP-ACM}} & \multicolumn{3}{r}{\textbf{DBLP-Scholar}} \\\hline
    
&& \textbf{expected} & \textbf{execution} & \textbf{RMSE}  & \textbf{expected} & \textbf{execution} & \textbf{RMSE}  & \textbf{expected} & \textbf{execution} & \textbf{RMSE}\\  
    
\multirow{3}{*}{\emph{PPJoin+}} &\textbf{exp}&7.33 & 14.45 & 2.78 & 8.36	& 14.56 & 2.43	& 177.02 &	124.88 & 8.02 \\

&\textbf{linear}&	8.37&	16.24&	3.28	&	7.45	&15.81	&2.97	&	222.55	&147.33&	9.48	\\

&\textbf{mixed}&	6.09&	13.45&	2.70	&	6.12	&16.83&	3.56	&	129.63	&149.82	&6.69\\\hline

\multirow{3}{*}{\emph{Ed-Join}} &\textbf{exp}&22.81	&27.33&	3.89	&	34.33&	36.84	&3.49	&	428.93&	324.79	&12.31\\

&\textbf{linear}	&17.99&26.04	& 2.60&	25.29	&35.85	&3.35		&354.97	&404.06	&9.65	\\

&\textbf{mixed}	&18.34	&26.45	&2.78		&27.68	&41.20	&3.54		&338.55	&339.31	&7.30	\\\hline

\textbf{Measures}& \textbf{Model}&\multicolumn{3}{r}{\textbf{MOVIES}} & \multicolumn{3}{r}{\textbf{VILLAGES}} && \textbf{AVERAGE}\\\hline

&&\textbf{expected} & \textbf{execution} & \textbf{RMSE}& \textbf{expected}& \textbf{execution} & \textbf{RMSE}\\

\multirow{3}{*}{\emph{PPJoin+}} &\textbf{exp}& 134.90& 146.39& 5.44 & 211.89 & 135.53& 9.36&\multirow{3}{*}{\emph{PPJoin+}} &\textbf{exp}&5.61\\

&\textbf{linear}&38.60 &33.10	&2.95	&	158.89&	131.64&	5.23&&\textbf{linear}&\textbf{4.78}\\

&\textbf{mixed}&48.45&	49.89	&3.17	&	214.15&	201.17&	8.13&&\textbf{mixed}&4.85\\\hline

\multirow{3}{*}{\emph{Ed-Join}} &\textbf{exp}& 59.57&	45.47&	3.76		&1,225.57	&1,556.04&	35.23&\multirow{3}{*}{\emph{Ed-Join}} &\textbf{exp}&11.74\\

&\textbf{linear}	&43.02	&44.46	& 3.52	&509.71	&294.35	&22.53	&&\textbf{linear}	&8.33\\

&\textbf{mixed}	&45.55	&43.26	&2.88		&377.02	&286.91	&10.89&&\textbf{mixed}&\textbf{5.48}\\\hline
\end{tabular}
}
\end{table}

To continue with $Q_2$, we conducted a set of experiments in order to observe how well each model could generalize among the different classes included in our evaluation data. Tables~\ref{tab:exp2Acm},~\ref{tab:exp2Amazon},~\ref{tab:exp2Scholar},~\ref{tab:exp2Movies} and ~\ref{tab:exp2Villages} present the results of training on one dataset and testing the resulting models on the set of the remaining classes.  The highest \emph{RMSE} error was achieved when both measures were tested using the \emph{exp} model in all datasets but VILLAGES. However, Table~\ref{tab:exp2Villages} shows that the fitting error when trained on VILLAGES is relatively low among all three models. 
Additionally, we observe that the \emph{exp} model's \emph{RMSE} increased exponentially as the quantity of the training data decreased, which constitutes this model as inadequate and unreliable for runtime approximations. By observing Tables~\ref{tab:exp2Acm} and ~\ref{tab:exp2Scholar}, we see that the \emph{RMSE} of the \emph{exp} model increased by 38 orders of magnitude for \emph{Ed-Join}. 

For both measures, the \emph{linear} model outperformed the other two models on average when trained on the Amazon-GP, DBLP-ACM and DBLP-Scholar datasets and achieved the lowest \emph{RMSE} when trained on MOVIES for \emph{Ed-Join} compared to \emph{exp} and \emph{mixed}. Both \emph{linear} and \emph{mixed} achieved minuscule approximation errors compared to \emph{exp}, but \emph{linear} was able to produce at least 35\% less \emph{RMSE} compared to \emph{mixed}. Therefore, we can answer $Q_2$ by stating that the \emph{linear}
model is the most suitable and sufficient model that can generalize among different classes.

\begin{table}[htb]
\centering
\caption{Average expected runtime, average execution time and root mean square error for training on Amazon-GP dataset and testing on DBLP-ACM, DBLP-Scholar, MOVIES and VILLAGES. All runtimes are presented in milliseconds.}

\label{tab:exp2Amazon}
\resizebox{1.0\textwidth}{!}
{
\begin{tabular}{rr|rrr|rrr|rrr} \hline

\textbf{Measures}& \textbf{Model}& \multicolumn{3}{r}{\textbf{DBLP-ACM}} & \multicolumn{3}{r}{\textbf{DBLP-Scholar}} & \multicolumn{2}{r}{\textbf{AVERAGE}}\\ \hline
    
&& \textbf{expected} & \textbf{execution} & \textbf{RMSE} &\textbf{expected} & \textbf{execution}&\textbf{RMSE}&\\  
    
\multirow{3}{*}{\emph{PPJoin+}}&\textbf{exp}&18.24&64.02&8.61&1.84E+17&1,609.71&	1.84E+16	&&&\\

&\textbf{linear}&25.42&87.68	&12.23&409.98&474.82	&20.59
&\multirow{3}{*}{\emph{PPJoin+}}&\multirow{2}{*}{\textbf{exp}}&\multirow{2}{*}{8.42E+35}\\

&\textbf{mixed}&	44.67&137.54&18.72&270.33&339.06&20.02
&&\multirow{2}{*}{\textbf{linear}}&\multirow{2}{*}{\textbf{24.68}} \\

\multirow{3}{*}{\emph{Ed-Join}}&	\textbf{exp}	&62.62&142.76&15.67 &5.34E+19&834.11&5.34E+18
&&\multirow{2}{*}{\textbf{mixed}}&\multirow{2}{*}{90.07}\\

&\textbf{linear}&	37.19&	131.68&	19.26&	663.07	&837.88&	27.30\\

&\textbf{mixed}&	38.36&140.25&16.87&	770.51&861.72&21.91\\\hline
\textbf{Measures}& \textbf{Model} & \multicolumn{3}{r}{\textbf{MOVIES}} & \multicolumn{3}{r}{\textbf{VILLAGES}} & \\ \hline

&& \textbf{expected} & \textbf{execution} & \textbf{RMSE} & \textbf{expected}& \textbf{execution} & \textbf{RMSE}&\\

\multirow{3}{*}{\emph{PPJoin+}}& \textbf{exp}& 8.79E+05 &	95.28&8.79E+04 &3.37E+37	&352.77	&3.37E+36&\\

&\textbf{linear}&133.06&202.34&11.32	&853.58	&331.61	&54.62
&\multirow{3}{*}{\emph{Ed-Join}}&\multirow{2}{*}{\textbf{exp}}&\multirow{2}{*}{8.43E+41}\\

&\textbf{mixed}&	136.17&98.58&6.37&3,507.19&360.03&315.15
&&\multirow{2}{*}{\textbf{linear}}&\multirow{2}{*}{\textbf{28.01}} \\

\multirow{3}{*}{\emph{Ed-Join}}&	\textbf{exp}&1.26E+07& 143.93&1.26E+06&9.75E+42&6,108.37	&9.75E+41
&&\multirow{2}{*}{\textbf{mixed}}&\multirow{2}{*}{54.49}\\

&\textbf{linear}&209.13&142.45&9.14&1,379.12&864.31&	56.32\\

&\textbf{mixed}&332.13&145.46&19.83&	7,258.82	&5,973.70&159.37\\\hline

\end{tabular}
}
\end{table}

\begin{table}[htb]
\centering
\caption{Average expected runtime, average execution time and root mean square error for training on DBLP-ACM dataset and testing on Amazon-GP, DBLP-Scholar, MOVIES and VILLAGES. All runtimes are presented in milliseconds.}
\label{tab:exp2Acm}
\resizebox{1.0\textwidth}{!}
{
\begin{tabular}{rr|rrr|rrr|rrr} \hline

\textbf{Measures}& \textbf{Model}& \multicolumn{3}{r}{\textbf{Amazon-GP}} & \multicolumn{3}{r}{\textbf{DBLP-Scholar}} & \multicolumn{2}{r}{\textbf{AVERAGE}}\\ \hline
    
&& \textbf{expected} & \textbf{execution} & \textbf{RMSE} &\textbf{expected} & \textbf{execution}&\textbf{RMSE}&\\  
    
\multirow{3}{*}{\emph{PPJoin+}}&\textbf{exp}&21.51	&61.69	&9.93	&	1.29E+16&	3,741.58&1.29E+15&&&\\

&\textbf{linear}&15.73&	46.13&	8.95		&346.71&	3,674.06&	341.87
&\multirow{3}{*}{\emph{PPJoin+}}&\multirow{2}{*}{\textbf{exp}}&\multirow{2}{*}{3.99E+15}\\

&\textbf{mixed}&	44.09	&120.62	&12.82&		534.41&	1,833.07	&139.71
&&\multirow{2}{*}{\textbf{linear}}&\multirow{2}{*}{\textbf{101.82}} \\

\multirow{3}{*}{\emph{Ed-Join}}&	\textbf{exp}	&85.53	&92.78	&8.02	&	2.82E+18&	888.50&	2.82E+17
&&\multirow{2}{*}{\textbf{mixed}}&\multirow{2}{*}{531.95}\\

&\textbf{linear}&56.95&	90.10 &7.91	&	950.61	&883.01	&25.97\\

&\textbf{mixed}&	58.29&	96.63	&8.48	&	1,472.52&	881.22&	63.72\\\hline
\textbf{Measures}& \textbf{Model} & \multicolumn{3}{r}{\textbf{MOVIES}} & \multicolumn{3}{r}{\textbf{VILLAGES}} & \\ \hline

&& \textbf{expected} & \textbf{execution} & \textbf{RMSE} & \textbf{expected}& \textbf{execution} & \textbf{RMSE}&\\

\multirow{3}{*}{\emph{PPJoin+}}&\textbf{exp}&8.05E+05	&108.16	&8.05E+04	&	1.47E+37 & 356.93&	1.47E+36 &\\

&\textbf{linear}&127.07	&132.62	&7.64	&	819.98&	368.86&	48.82
&\multirow{3}{*}{\emph{Ed-Join}}&\multirow{2}{*}{\textbf{exp}}&\multirow{2}{*}{9.3E+42}\\

&\textbf{mixed}&159.36	&120.74&	8.92		&2.14E+04	&1,783.72&	1,966.38
&&\multirow{2}{*}{\textbf{linear}}&\multirow{2}{*}{\textbf{53.95}} \\

\multirow{3}{*}{\emph{Ed-Join}}&	\textbf{exp}&3.58E+07&	156.97 & 3.58E+06		&3.72E+44	&6,329.54	&3.72E+43
&&\multirow{2}{*}{\textbf{mixed}}&\multirow{2}{*}{1,105.15}\\

&\textbf{linear}&373.99&	156.72	&23.23&		2,440.64&	870.15&	158.72\\

&\textbf{mixed}&1,246.20 &155.42&	109.39	&	4.87E+04& 	6,411.76	&4,239.01\\\hline

\end{tabular}
}
\end{table}

\begin{table}[htb]
\centering
\caption{Average expected runtime, average execution time and root mean square error for training on DBLP-Scholar dataset and testing on Amazon-GP, DBLP-ACM, MOVIES and VILLAGES. All runtimes are presented in milliseconds.}
\label{tab:exp2Scholar}
\resizebox{1.0\textwidth}{!}
{
\begin{tabular}{rr|rrr|rrr|rrr} \hline

\textbf{Measures}& \textbf{Model}& \multicolumn{3}{r}{\textbf{Amazon-GP}} & \multicolumn{3}{r}{\textbf{DBLP-ACM}} & \multicolumn{2}{r}{\textbf{AVERAGE}}\\ \hline
    
&& \textbf{expected} & \textbf{execution} & \textbf{RMSE} &\textbf{expected} & \textbf{execution}&\textbf{RMSE}&\\  
    
\multirow{3}{*}{\emph{PPJoin+}}&\textbf{exp}&79.32&	65.28&	8.03&47.42&	69.70&8.74 &&&\\

&\textbf{linear}&-364.95	&38.47	&40.61	&	173.40 &88.48	&15.39
&\multirow{3}{*}{\emph{PPJoin+}}&\multirow{2}{*}{\textbf{exp}}&\multirow{2}{*}{4.56E+04}\\

&\textbf{mixed}&	-41.05	&50.27&	11.00&-148.99&	88.14&	26.03
&&\multirow{2}{*}{\textbf{linear}}&\multirow{2}{*}{\textbf{85.07}} \\

\multirow{3}{*}{\emph{Ed-Join}}&	\textbf{exp}	&113.56	&80.90 &8.67		&113.43	&139.78	&16.74
&&\multirow{2}{*}{\textbf{mixed}}&\multirow{2}{*}{427.54}\\

&\textbf{linear}&44.49&	79.97	&10.67	&	37.70 &144.33	&22.36\\

&\textbf{mixed}&40.13	&73.76&8.98	&	40.94&	141.33	&18.84\\\hline
\textbf{Measures}& \textbf{Model} & \multicolumn{3}{r}{\textbf{MOVIES}} & \multicolumn{3}{r}{\textbf{VILLAGES}} & \\ \hline

&& \textbf{expected} & \textbf{execution} & \textbf{RMSE} & \textbf{expected}& \textbf{execution} & \textbf{RMSE}&\\

\multirow{3}{*}{\emph{PPJoin+}}&\textbf{exp}&110.41&	94.69&	6.31	&	1.82E+06&	1,546.07	&1.82E+05&\\

&\textbf{linear}&394.74	&104.19&	29.99	&	3,158.25&	621.84	&254.30
&\multirow{3}{*}{\emph{Ed-Join}}&\multirow{2}{*}{\textbf{exp}}&\multirow{2}{*}{1.10E+04}\\

&\textbf{mixed}&66.96	&85.61&	6.76		&1.82E+04&	1,591.24&	1,666.38
&&\multirow{2}{*}{\textbf{linear}}&\multirow{2}{*}{\textbf{54.57}} \\

\multirow{3}{*}{\emph{Ed-Join}}&	\textbf{exp}&341.02	&128.33	&22.66	&	4.46E+05	&6,069.92&	4.41E+04
&&\multirow{2}{*}{\textbf{mixed}}&\multirow{2}{*}{82.52}\\

&\textbf{linear}&360.47	&127.76	&24.51&		2,418.34	&818.14&	160.73\\

&\textbf{mixed}&280.77	&125.19&	16.86	&	3,670.31 &	820.85	&285.43\\\hline

\end{tabular}
}
\end{table}

\begin{table}[htb]
\centering
\caption{Average expected runtime, average execution time and root mean square error for training on MOVIES dataset and testing on Amazon-GP, DBLP-ACM, DBLP-Scholar and VILLAGES. All runtimes are presented in milliseconds.}
\label{tab:exp2Movies}
\resizebox{1.0\textwidth}{!}
{
\begin{tabular}{rr|rrr|rrr|rrr} \hline

\textbf{Measures}& \textbf{Model}& \multicolumn{3}{r}{\textbf{Amazon-GP}} & \multicolumn{3}{r}{\textbf{DBLP-ACM}} & \multicolumn{2}{r}{\textbf{AVERAGE}}\\ \hline
    
&& \textbf{expected} & \textbf{execution} & \textbf{RMSE} &\textbf{expected} & \textbf{execution}&\textbf{RMSE}&\\  
    
\multirow{3}{*}{\emph{PPJoin+}}&\textbf{exp}&19.53&	71.55	&7.89	&	46.89	&127.70&15.90&&&\\

&\textbf{linear}&-45.99&	42.58&	10.51	&	57.73	&120.70&23.93
&\multirow{3}{*}{\emph{PPJoin+}}&\multirow{2}{*}{\textbf{exp}}&\multirow{2}{*}{8.42E+06}\\

&\textbf{mixed}&	16.97	&39.64	&5.84	&	17.43	&66.84	&9.77
&&\multirow{2}{*}{\textbf{linear}}&\multirow{2}{*}{51.34} \\

\multirow{3}{*}{\emph{Ed-Join}}&	\textbf{exp}	&15.57&	80.95&	9.37	&	16.24	&135.66	&17.93
&&\multirow{2}{*}{\textbf{mixed}}&\multirow{2}{*}{\textbf{37.99}}\\

&\textbf{linear}&1.71&	84.53&	10.82&		3.56&138.18	&19.89	\\

&\textbf{mixed}&4.33&85.70 &10.95	&	6.99	&140.99	&19.65	\\\hline
\textbf{Measures}& \textbf{Model} & \multicolumn{3}{r}{\textbf{DBLP-Scholar}} & \multicolumn{3}{r}{\textbf{VILLAGES}} & \\ \hline

&& \textbf{expected} & \textbf{execution} & \textbf{RMSE} & \textbf{expected}& \textbf{execution} & \textbf{RMSE}&\\

\multirow{3}{*}{\emph{PPJoin+}}&\textbf{exp}&3,636.56&	318.89&	332.11	&3.37E+08	&634.17	&3.37E+07&\\

&\textbf{linear}&372.82	&1,315.61&	102.21	&1,064.96&	389.93&	68.69
&\multirow{3}{*}{\emph{Ed-Join}}&\multirow{2}{*}{\textbf{exp}}&\multirow{2}{*}{1.46E+06}\\

&\textbf{mixed}&75.49&	702.11	&67.82		&989.17	&311.60&	68.54
&&\multirow{2}{*}{\textbf{linear}}&\multirow{2}{*}{\textbf{25.91}} \\

\multirow{3}{*}{\emph{Ed-Join}}&	\textbf{exp}&4,060.80 &811.77	&325.48&		5.85E+07	&767.66	&5.85E+06
&&\multirow{2}{*}{\textbf{mixed}}&\multirow{2}{*}{42.85}\\

&\textbf{linear}&259.61	&805.29	&57.92&696.29	&753.35	&15.04\\

&\textbf{mixed}&178.93&	796.16&	65.09&1,522.63	&777	.00&75.74\\\hline

\end{tabular}
}
\end{table}

\begin{table}[htb]
\centering
\caption{Average expected runtime, average execution time and root mean square error for training on VILLAGES dataset and testing on Amazon-GP, DBLP-ACM, DBLP-Scholar and MOVIES. All runtimes are presented in milliseconds.}
\label{tab:exp2Villages}
\resizebox{1.0\textwidth}{!}
{
\begin{tabular}{rr|rrr|rrr|rrr} \hline

\textbf{Measures}& \textbf{Model}& \multicolumn{3}{r}{\textbf{Amazon-GP}} & \multicolumn{3}{r}{\textbf{DBLP-ACM}} & \multicolumn{2}{r}{\textbf{AVERAGE}}\\ \hline
    
&&\textbf{expected} & \textbf{execution} & \textbf{RMSE} &\textbf{expected} & \textbf{execution}&\textbf{RMSE}&\\  
    
\multirow{3}{*}{\emph{PPJoin+}}&\textbf{exp}&93.41&	67.44&5.08&35.07&62.53&8.36 &&&\\

&\textbf{linear}&-192.27&24.57&21.87	&-133.03&61.10&21.09
&\multirow{3}{*}{\emph{PPJoin+}}&\multirow{2}{*}{\textbf{exp}}&\multirow{2}{*}{\textbf{10.16}}\\

&\textbf{mixed}&	16.37	&32.66	&3.40&41.57	&61.83	&9.20
&&\multirow{2}{*}{\textbf{linear}}&\multirow{2}{*}{22.91} \\

\multirow{3}{*}{\emph{Ed-Join}}&	\textbf{exp}	&68.00	&53.36&4.50&326.05	&143.84	&26.53
&&\multirow{2}{*}{\textbf{mixed}}&\multirow{2}{*}{30.59}\\

&\textbf{linear}&-123.44	&55.03	&18.20&-677.4&	133.63&	82.36\\

&\textbf{mixed}&231.61&50.46&	18.51&136.49	&139.30&	15.95\\\hline
\textbf{Measures}& \textbf{Model} & \multicolumn{3}{r}{\textbf{DBLP-Scholar}} & \multicolumn{3}{r}{\textbf{MOVIES}} & \\ \hline

&& \textbf{expected} & \textbf{execution} & \textbf{RMSE} & \textbf{expected}& \textbf{execution} & \textbf{RMSE}&\\

\multirow{3}{*}{\emph{PPJoin+}}&\textbf{exp}&92.10&272.40	&21.78&56.74	&82.92	&5.43&\\

&\textbf{linear}&-39.98	&277.56	&34.10&-54.33&84.08	&14.57
&\multirow{3}{*}{\emph{Ed-Join}}&\multirow{2}{*}{\textbf{exp}}&\multirow{2}{*}{\textbf{21.59}}\\

&\textbf{mixed}&84.22&451.80	&40.04	&-26.91	&651.50	&69.71
&&\multirow{2}{*}{\textbf{linear}}&\multirow{2}{*}{54.56} \\

\multirow{3}{*}{\emph{Ed-Join}}&	\textbf{exp}&316.66	&784.7&49.85	&138.63	&114.50	&5.46
&&\multirow{2}{*}{\textbf{mixed}}&\multirow{2}{*}{32.75}\\

&\textbf{linear}&159.75	&753.00&	61.23&-438.84	&122.89&	56.44\\

&\textbf{mixed}&1,737.75	&945.09	&81.94&255.96	&116.42&	14.61\\\hline

\end{tabular}
}
\end{table}

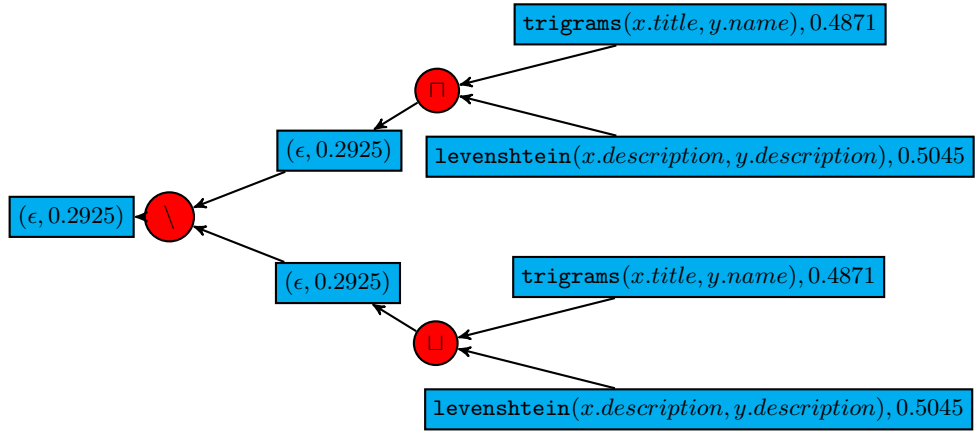
\begin{figure}[htb]
	\centering
\begin{tikzpicture}[scale = 1, every node/.style={scale=1}]
	  \begin{scope}[thick,
			node distance=.6cm,on grid,>=stealth',
			filter/.style={rectangle,draw,fill=cyan},
			op/.style={circle,draw,fill=red}]
		\node [filter](filter)[]{$(\epsilon, 0.2925)$};		
		\node [op](minus)[right =of filter,xshift=0.7cm]{$\backslash$} edge [->] (filter);
		\node [filter](filter1)[above=of minus.west,xshift=2.6cm]	{$(\epsilon, 0.2925)$} edge [->] (minus);
		\node [op](intersection)[right =of filter1,xshift=0.7cm,yshift=0.8cm]{$\sqcap$} edge [->] (filter1);
		\node [filter] (label1)[above =of intersection.west,xshift=3.8cm]{$\texttt{trigrams}(x.title,y.name), 0.4871$} edge [->] (intersection); 	
		\node [filter] (label2)[below =of intersection.west,xshift=3.8cm]{$\texttt{levenshtein}(x.description,y.description), 0.5045$} edge [->] (intersection); 
		\node [filter](filter2)[below=of minus.east,xshift=1.9cm]	{$(\epsilon, 0.2925)$} edge [->] (minus);
		\node [op](union)[right =of filter2,xshift=0.7cm,yshift=-0.8cm]{$\sqcup$} edge [->] (filter2);
		\node [filter] (label3)[above =of union.west,xshift=3.8cm]{$\texttt{trigrams}(x.title,y.name), 0.4871$} edge [->] (union); 	
		\node [filter] (label4)[below =of union.west,xshift=3.8cm]{$\texttt{levenshtein}(x.description,y.description), 0.5045$} edge [->] (union);
	  \end{scope}
	\end{tikzpicture}
	\caption{Plan returned from \HELIOS using the \emph{exp} model.}
	\label{fig:exp}
\end{figure}

\begin{figure}[htb]
	\centering
\begin{tikzpicture}[scale = 1, every node/.style={scale=1}]
	  \begin{scope}[thick,
			node distance=.9cm,on grid,>=stealth',
			filter/.style={rectangle,draw,fill=cyan},
			op/.style={circle,draw,fill=red}]
		\node [filter](filter0)[]{$(\epsilon, 0.2925)$};	
		\node [op](minus)[below =of filter0,]{$\backslash$} edge [->] (filter);
		
		\node [filter](filter)[below =of minus]{$(\epsilon, 0.2925)$} edge [->] (minus);		
		
		\node [filter](filterIntersection)[below =of filter]{$\texttt{levenshtein}(x.description,y.description), 0.5045$} edge [->] (filter);
				\node [filter](filter1)[below=of filterIntersection]	{$\texttt{trigrams}(x.title,y.name), 0.4871$} edge [->] (filterIntersection);

		\node [filter](filter2)[right=of minus.east]	{$(\epsilon, 0.2925)$} edge [->] (minus);
		\node [op](union)[right =of filter2,xshift=0.7cm]{$\sqcup$} edge [->] (filter2);
		\node [filter] (label3)[above =of union.west,xshift=0.3cm]{$\texttt{trigrams}(x.title,y.name), 0.4871$} edge [->] (union); 	
		\node [filter] (label4)[below =of union.west,xshift=1.6cm,yshift=0.3cm]{$\texttt{levenshtein}(x.description,y.description), 0.5045$} edge [->] (union);
		
	  \end{scope}
	\end{tikzpicture}		
\caption{Plan returned from \HELIOS using the \emph{linear} and \emph{mixed} model.}\label{fig:linear}
\end{figure}
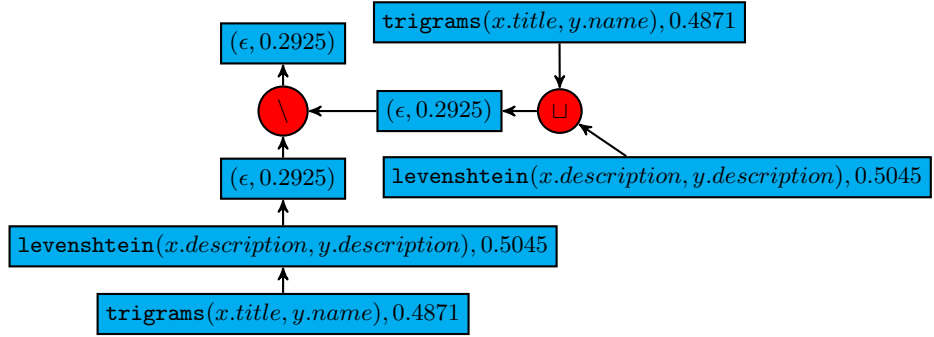

For our last question, we tested the performance of the different models when trained on a bigger and more diverse dataset. Table~\ref{tab:exp3} shows the results of our evaluation, where each model was trained on DBpedia english labels and tested on the the four evaluation datasets.  The \emph{linear} model error was 1 order of magnitude less than the \emph{RMSE} obtained by \emph{exp} and 3 orders of magnitude less compared to the \emph{mixed} error. In all four datasets, the \emph{mixed} model produced the highest \emph{RMSE}. 
For the VILLAGES dataset, the \emph{mixed} model's error was 1916 and 214 times higher compared to \emph{linear} and \emph{exp} resp. Figs.~\ref{fig:exp} and ~\ref{fig:linear} present the plans produces by \HELIOS for the LS \texttt{MINUS(AND(levenshtein(x.description,y.description)$|$0.5045,trigrams(\\x.title, y.name)$|$0.4871)$|$0.2925,OR(levenshtein(x.description,y.descri\\ption)$|$0.5045,trigrams(x.title, y.name)$|$0.4871)$|$ 0.2925)$>=$0.2925} of the Amazon-GP dataset, if the planner used the \emph{exp} model and the \emph{linear} or the \emph{mixed} model resp. For the child LS \texttt{AND(levenshtein(x.description,y.descri\\ption)$|$0.5045,trigrams(x.title, y.name)$|$0.4871)$|$0.2925}, the \emph{linear} and the \emph{mixed} model chose to execute only \texttt{trigrams(x.title, y.name)$|$0.4871)} and use the other child as a filter. Moreover, the plan retrieved by using the \emph{exp} model for runtime approximations aims to execute both children LSs, which results into an overhead in the execution of the LS. It is obvious that the \emph{linear} model achieved by far the lowest \emph{RMSE} on average compared to the other two models, which concludes the answer to $Q_3$.

\begin{table}[htb]
\centering
\caption{Average expected runtime, average execution time and root mean square error for training on DBPedia english labels and testing on Amazon-GP, DBLP-ACM, MOVIES and VILLAGES. All runtimes are presented in milliseconds.}
\label{tab:exp3}
\resizebox{1.0\textwidth}{!}
{
\begin{tabular}{r|rrr|rrr|rrr} \hline

\textbf{Model}&\multicolumn{3}{r}{\textbf{Amazon-GP}} & \multicolumn{3}{r}{\textbf{DBLP-ACM}} & \multicolumn{2}{r}{\textbf{AVERAGE}}\\ \hline
    
&\textbf{expected} & \textbf{execution} & \textbf{RMSE} 
&\textbf{expected} & \textbf{execution}&\textbf{RMSE}&\\  
    
\textbf{exp}&5,242.09	&3,618.99	&3,164.86	&	308.14&	365.46&	126.42
\\

\textbf{linear}&300.51	&3,043.97&	966.99	&	8.07&	361.53	&192.12
&\multirow{3}{*}{\textbf{exp}}&\multirow{3}{*}{4,577.58} \\

\textbf{mixed}&-7.27E+06	&4,512.82&	6.78E+05	&	-7.26E+04&	310.49&	4.38E+04	
&\multirow{3}{*}{\textbf{linear}}&\multirow{3}{*}{\textbf{512.35}}\\

\textbf{Model}&\multicolumn{3}{r}{\textbf{Amazon-GP}} & \multicolumn{3}{r}{\textbf{DBLP-ACM}} &\multirow{3}{*}{\textbf{mixed}}&\multirow{3}{*}{9.82E+05}\\  

\textbf{expected} & \textbf{execution} & \textbf{RMSE} &\textbf{expected} & \textbf{execution}&\textbf{RMSE}&\\  
    
\textbf{exp}&584.27&	1,061.67	&160.05		&4.61E+04	&3,775.54	&1.48E+04 
\\

\textbf{linear}&323.04	&995.04	&258.55		&2,626.41&	3,832.52	&631.72
\\

\textbf{mixed}&-3,417.80 &1,600.81	&2,042.45	&	7.15E+06 	&3,891.05&	3.20E+06 	
\\ \hline

\end{tabular}
}
\end{table}
%
%
%
%
%
%
%

\section{Related Work}
The task of efficient query execution in database systems is similar to the task of execution optimization using runtime approximations in LD frameworks. Efficient and scalable data management has been of central importance in database systems \cite{Graefe}. Over the past few years, there has been an extensive work on query optimization in databases that is based on statistical information about relations and intermediate results \cite{SIL+06}. The author of \cite{CHAU98} gives an analytic overview regarding the procedure of query optimization and the different approaches used at each step of the process. 

A novel approach in this field was presented by \cite{Ioannidis}, in which the proposed approach introduced the concept of parametric query optimization. In this work, the authors provided the necessary formalization of the aforementioned concept and conducted a set of experiments using the buffer size as parameter. In order to minimize the total cost of generating all possible alternative execution plans, they used a set of randomized algorithms. On a similar manner, the authors of \cite{multi} introduced the idea of Multi-Objective Parametric query optimization (MPQ), where the cost of plan is associated with multiple cost functions and each cost function is associated with various parameters. Their experimental results showed however that the MPQ method performs an exhaustive search of the solution space which addresses this approach computationally inefficient. 

Another set of approaches in the field of query optimization have focused on creating dynamic execution plans. Dynamic planning is based on the idea that the execution engine of a framework knows more than the planner itself. Therefore, information generated
by the execution engine is used to re-evaluate the plans generated by the optimizer. There has been a vast amount of approaches towards dynamic query optimization such as query scrambling for initial delays \cite{Urhan}, dynamic planning in compile-time \cite{Cole}, adaptive query operators \cite{ives} and re-ordering of operators \cite{Avnur}. 

Moreover, the problem addressed in this work focus on identifying scalable and time-efficient solutions towards LD. A large number of frameworks were developed to assist this issue, such as SILK \cite{ISE+11}, \LIMES \cite{NGON12c}, KnoFuss \cite{NIK+12} and Zhishi.links \cite{XIN+11}. SILK and KnoFuss implement blocking approaches in order to achieve efficient linking between resources. SILK framework incorporates a rough index pre-match, whereas KnoFuss blocking technique is highly influenced from databases systems techniques. To this end, the only LD framework that provides both theoretical and practical guarantees towards scalable and accurate LD is \LIMES. As we mentioned throughout this work,\LIMES execution strategy incorporates the \HELIOS planner ~\cite{helios} which is (to the best of our knowledge) the first execution optimizer in LD. \HELIOS is able to provide accurate runtime approximations, which we have extended in this work, and is able to find the least costly execution plan for a LS, consuming a minute portion of the overall execution runtime.

\section{Conclusion}
In this paper, we studied approximation functions that allow predicting the runtime of link specifications. We showed that on average, linear models are indeed the approach to chose to this end as they seem to overfit the least. Still, mixed models also perform in a satisfactory manner. Exponential models either fit very well or not at all and are thus not to be used. In future work, we will study further models for the evaluation of  runtime and improve upon existing planning mechanisms for the declarative LD. In particular, we will consider other features when approximation runtimes, e.g., the distribution of characters in the strings to compare.


\end{document}